\documentclass[sigconf]{acmart}

\makeatletter
\def\@ACM@checkaffil{
    \if@ACM@instpresent\else
    \ClassWarningNoLine{\@classname}{No institution present for an affiliation}%
    \fi
    \if@ACM@citypresent\else
    \ClassWarningNoLine{\@classname}{No city present for an affiliation}%
    \fi
    \if@ACM@countrypresent\else
        \ClassWarningNoLine{\@classname}{No country present for an affiliation}%
    \fi
}
\makeatother

\usepackage{microtype}
\usepackage{graphicx}
\usepackage{booktabs} 
\usepackage{amsfonts}
\usepackage{xcolor}

\usepackage{caption}
\usepackage{subcaption}
\usepackage{tikz}
\usepackage{multirow}
\usepackage{array}
\usepackage{enumitem}

\usepackage[subtle]{savetrees}
\newcommand{\G}{\mathcal{G}}

\newcommand{\M}{\mathbf{M}}
\newcommand{\R}{\mathbb{R}}

\AtBeginDocument{%
  \providecommand\BibTeX{{%
    \normalfont B\kern-0.5em{\scshape i\kern-0.25em b}\kern-0.8em\TeX}}}

\copyrightyear{2024}
\acmYear{2024}
\setcopyright{rightsretained}
\acmConference[CIKM '24]{Proceedings of the 33rd ACM International Conference on Information and Knowledge Management}{October 21--25, 2024}{Boise, ID, USA}
\acmBooktitle{Proceedings of the 33rd ACM International Conference on Information and Knowledge Management (CIKM '24), October 21--25, 2024, Boise, ID, USA}
\acmDOI{10.1145/3627673.3680021}
\acmISBN{979-8-4007-0436-9/24/10}
\begin{document}



\title[GraphScale: A Framework to Enable Machine Learning over Billion-node Graphs]{GraphScale: A Framework to Enable Machine Learning over \texorpdfstring{\\}{ }Billion-node Graphs}


\author{Vipul Gupta}
\affiliation{
\institution{Bytedance}
\city{San Jose}
  \country{USA}
}

\author{Xin Chen}
\affiliation{
\institution{Bytedance}
\city{San Jose}
  \country{USA}
}

\author{Ruoyun Huang}
\affiliation{
\institution{Bytedance, USA}
\city{Seattle}
  \country{USA}
}

\author{Fanlong Meng}
\affiliation{
\institution{Bytedance}
\city{San Jose}
  \country{USA}
}

\author{Jianjun Chen}
\affiliation{
\institution{Bytedance}
\city{San Jose}
  \country{USA}
}


\author{Yujun Yan}
\affiliation{%
  \institution{Dartmouth College}
  \city{Hanover}
  \country{USA}
}

\renewcommand{\shortauthors}{Vipul Gupta et al.}

\begin{abstract}
Graph Neural Networks (GNNs) have emerged as powerful tools for supervised machine learning over graph-structured data, while sampling-based node representation learning is widely utilized in unsupervised learning. However, scalability remains a major challenge in both supervised and unsupervised learning for large graphs (e.g., those with over 1 billion nodes). The scalability bottleneck largely stems from the mini-batch sampling phase in GNNs and the random walk sampling phase in unsupervised methods. These processes often require storing features or embeddings in memory. In the context of distributed training, they require frequent, inefficient random access to data stored across different workers. Such repeated inter-worker communication for each mini-batch leads to high communication overhead and computational inefficiency. 

We propose GraphScale, a unified framework for both supervised and unsupervised learning to store and process large graph data distributedly. The key insight in our design is the separation of workers who store data and those who perform the training. This separation allows us to decouple computing and storage in graph training, thus effectively building a pipeline where data fetching and data computation can overlap asynchronously. Our experiments show that GraphScale outperforms state-of-the-art methods for distributed training of both GNNs and node embeddings. We evaluate GraphScale both on public and proprietary graph datasets and observe a reduction of at least $40\%$ in end-to-end training times compared to popular distributed frameworks, without any loss in performance. While most existing methods don't support billion-node graphs for training node embeddings, GraphScale is currently deployed in production at TikTok enabling efficient learning over such large graphs.
\end{abstract}




\maketitle

\vspace{-2mm}

\section{Introduction}
\label{intro}

Graph learning is a powerful way to generalize traditional deep learning methods onto graph-structured data that contains interactions (edges) between individual units (nodes). Despite the promising performance of machine learning methods that learn directly from graph data \cite{gnn_review_book, gcn}, their adaption at modern web companies has been limited due to the sheer scale \cite{pbg_biggraph}. 
For example, the social network graph at Facebook includes over two billion
user nodes and over a trillion edges \cite{fb_graph_scale}, the users-products graph at Alibaba consists of more than a billion users and two billion items \cite{wang_alibaba}, and the user-to-item graph at Pinterest includes at least 2 billion entities and over 17 billion edges \cite{pinterest_embedding}. At Bytedance, we frequently observe graphs with over a billion nodes on data obtained from the social media platform TikTok.

Machine learning on graphs can be broadly classified into two categories. The {\bf first} is supervised learning, where the training process ingests the node features, edge features, graph structure, etc., and the labels for the supervised learning task at hand, for example, node classification and link prediction. Graph Neural Networks (GNNs) have become the de-facto way of supervised learning on graphs, popularized by methods such as Graph Convolution Network (GCN) \cite{gcn} and GraphSAGE \cite{graphsage}. GNNs have been applied to a broad range of applications, such as recommendation systems \cite{recommender_gcn} and drug discovery \cite{drug_discovery}, to obtain state-of-the-art results.

\begin{figure}[t]
\centering
\includegraphics[width=0.3\textwidth]{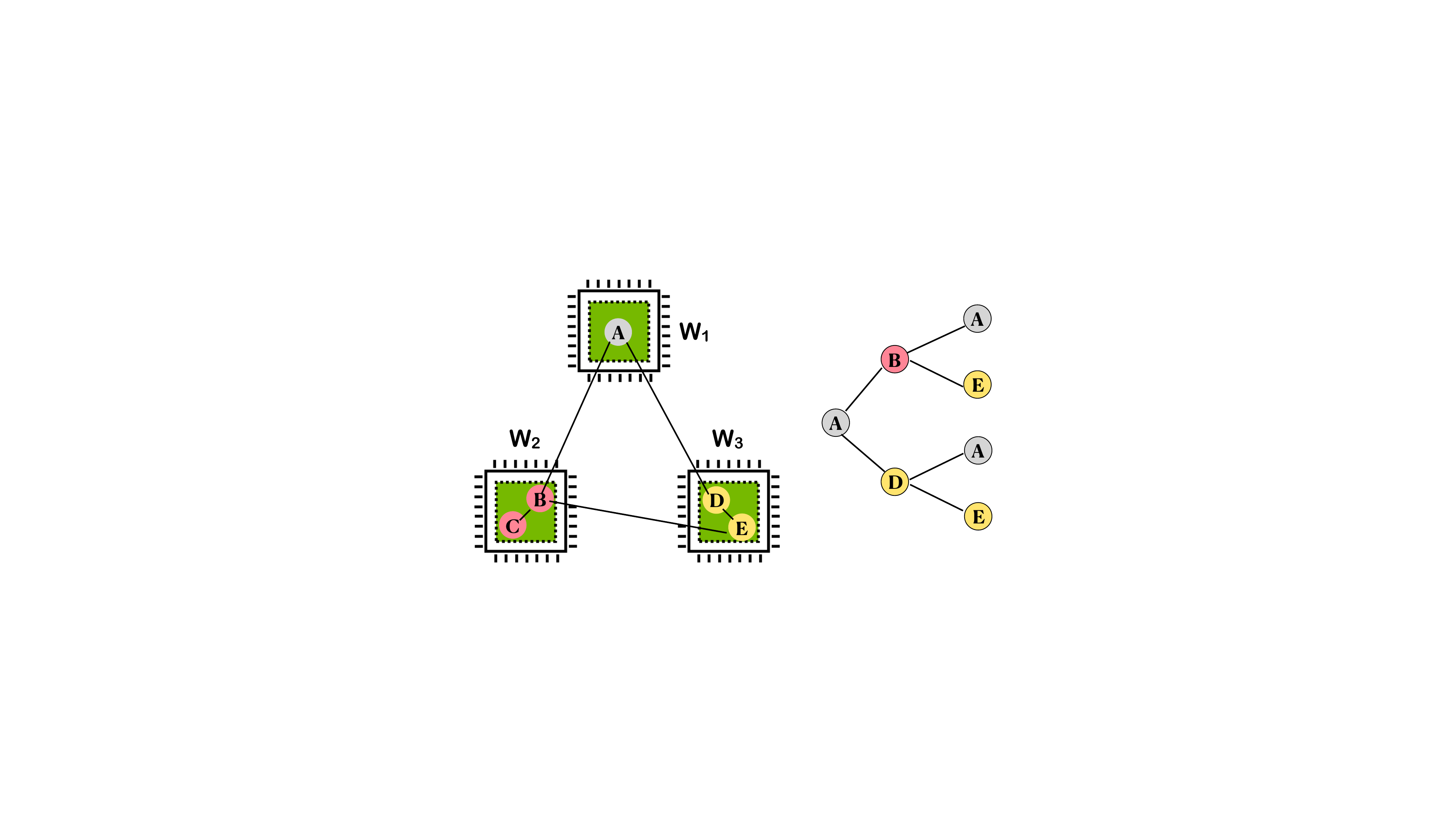}
\caption{\footnotesize Traditional distributed GNN training of a five-node graph with 3 workers ($W_1$, $W_2$ and $W_3$) in frameworks like GraphLearn and DistDGL. On the right, we show the computation graph to update node A with a fanout of [2,2], which requires 5 cross-device visits to fetch neighbor data for aggregation (2 and 1 visits required at nodes B and D, respectively, in the second hop, and 2 visits required at node A in the first hop), causing extensive communication overhead. 
}
\label{fig:gnn_comm_bottleneck}
\end{figure}

One critical bottleneck for mini-batch training of GNNs is the sampling phase, where for each vertex in the mini-batch, the sampler samples and fetches features of a subset of its neighboring nodes (and/or corresponding edges). The sampling phase can take a significantly longer time than the training phase due to a large amount of random data access 
and remote feature fetching, especially during distributed training of large graphs \cite{bytegnn, gns_da, mlsys_gnn_sampling}. This is illustrated through an example in Fig. \ref{fig:gnn_comm_bottleneck}, where we distribute a 5-node graph across three workers and the fanout for GNN training is [2,2]\footnote{A fanout of $[F_1, F_2, ..., F_h]$ in GNN training specifies that in the $i$-th layer of aggregation, each node samples information from up to $F_i$ neighbors for all $i\in[1,h]$. This hyperparameter controls the breadth of neighborhood sampling at each layer.}. In some cases, the incurred communication cost may account for $80\%$ or more of the training time \cite{sancus, p3, dgcl}.

GraphScale addresses this issue by decoupling feature fetching from the graph sampling process, which involves separating the computation and storage elements by using different workers to do the computation and storage. This separation allows overlapping feature fetching with computation. 
Also, each computation worker (referred to as a trainer) uses the full topology information to streamline the feature fetching process. This reduces the number of required feature requests, typically needing just one request per trainer for each mini-batch. 

The {\bf second} broad category is unsupervised learning to train graph embedding vectors.
Working with graph data directly, especially for the purposes of machine learning, is complex, and thus, a common technique is to use graph embedding methods to create vector representations for each node.
Node embedding methods are a way to convert graph data into a more computationally tractable form to allow them as an input to a variety of machine learning tasks \cite{hamilton2017representation}.
The representations can then serve as useful features for downstream tasks such as recommender systems \cite{wang_alibaba}, link prediction \cite{deepwalk}, predicting drug interactions \cite{bio_zitnik}, community detection \cite{community_detection}, etc. Hence, such methods are getting significant traction in both academia and industry \cite{embedding_framework}.

\begin{figure}[t]
\centering
\includegraphics[width=0.32\textwidth]{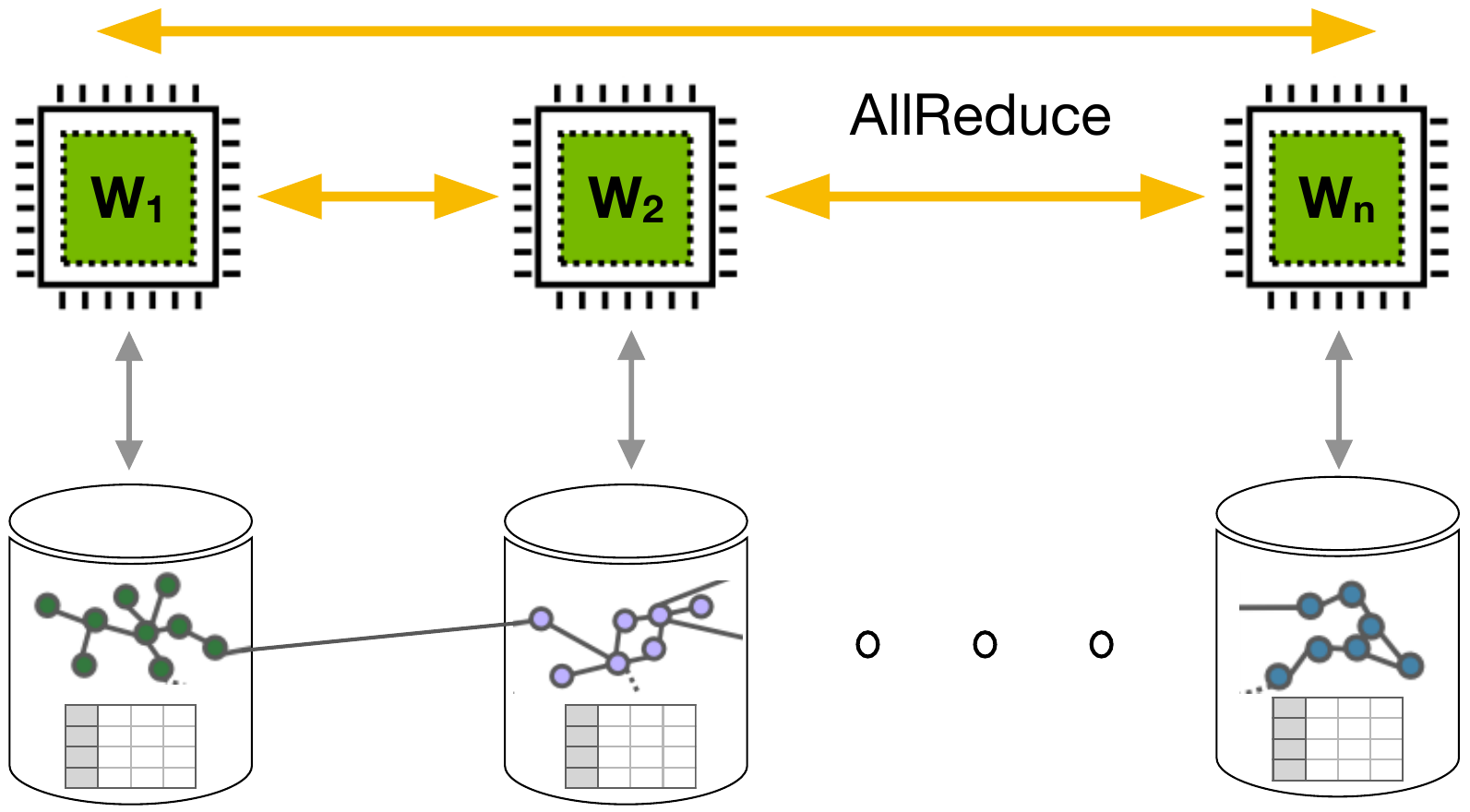}
\caption{\footnotesize Traditional data parallel training of node embedding matrix where each trainer stores one graph partition and the entire embedding matrix, which can lead to memory bottlenecks. 
Further, high communication cost is incurred during allreduce while averaging gradients in each iteration.}
\label{fig:graph_dp}
\end{figure}

One limitation of training node embeddings for large graphs is the size of the model\footnote{In this paper, we use the term ``model'' interchangeably with the embeddings matrix in the context of unsupervised learning.}. For example, the node embedding matrix for a graph with one billion nodes and an embedding dimension of 256 would require 512 GB of memory in 16-bit precision floats\footnote{Note that learning the node embeddings happens only for transductive methods. The embedding vectors are not trained for inductive methods, such as unsupervised Graphsage \cite{graphsage}. Instead, a neural network is trained to output embedding vectors for graph nodes based on their features and neighborhood subgraphs. However, such methods are not very good at capturing the structural information of the graph compared to transductive methods \cite{khosla2019comparative}.}. Traditional distributed training methods that use data-parallelism hold the entire model in memory. 
In this case, the data (that is, the graph that contains node and edge connections) is partitioned into subgraphs using partitioning schemes, e.g., Metis \cite{metis}, and each worker does mini-batch training from its own subgraph (as illustrated in Fig. \ref{fig:graph_dp}). Such data-parallel training of node embeddings has two primary bottlenecks: 1) Requires large memory to store the model (as well as gradient and momentum) data during training, and 2) High communication cost while averaging gradients after every iteration due to the size of the model. Such methods are infeasible or highly inefficient for large graphs like the ones we observe at TikTok.

To ameliorate these bottlenecks, we propose hybrid (that is, both data and model) parallelism for training node embeddings as a part of the GraphScale framework. 
Here, we divide the graph into disjoint subgraphs across multiple trainers, as well as the model (that is, the embedding matrix) across storage workers. This allows us to scale both computation and storage independently with the size of the graph. Further, model parallelism enables each worker to handle only a subset of the model, reducing the need for full gradient communication per iteration and offering significant communication savings over data parallelism alone.

This paper proposes a framework called GraphScale to address and mitigate key bottlenecks for graph learning.
The key features of GraphScale are:
\begin{itemize}[leftmargin=*]
\setlength{\parskip}{0pt}
\setlength{\itemsep}{0pt plus 0.2pt}

    \item {\bf Fast feature-fetching in GNN training}: 
    GraphScale optimizes distributed GNN training by decoupling feature fetching from graph sampling, using distinct workers for computation and storage. This reduces feature requests and allows overlapping fetching with computation to alleviate communication bottlenecks.
    \item {\bf Efficient node embedding training}: 
    GraphScale uses hybrid parallelism for training node embeddings at scale. It uses data parallelism to divide computation between workers and model parallelism 
    to reduce communication and storage requirements of the large node embedding matrix. 
    \item {\bf Serverless}: 
    GraphScale, leveraging the capabilities of Ray--a framework designed for distributed serverless computation--offers significant advantages such as elasticity in resource allocation and failure tolerance \cite{ray}. 
    With Ray, resources can be provisioned in real-time, and the complexities of managing them are abstracted away by Ray's user-friendly API. 
    This approach ensures that users of GraphScale can focus on their training tasks, while Ray efficiently manages the underlying computational resources.
    \item {\bf General Framework}:  GraphScale is a unified framework that works on both supervised and unsupervised learning methods and is agnostic to the type of models and algorithms used (such as GraphSage and GCN for supervised learning and DeepWalk, Node2Vec and LINE for unsupervised learning). It is also backend agnostic and can work with multiple graph frameworks such as DGL, PyTorch-geometric, and GraphLearn.
\end{itemize}



\section{Preliminaries}
\label{sec:priliminaries}

In this section, we review the basics of GNN and node embedding training.
We use the following notation throughout the paper. We want to perform learning over a graph $\G = (V, E)$ with $N= |V|$ nodes and $|E|$ edges. 
Each node has a feature vector of dimension $f$, and during supervised learning, the objective is to train a GNN based on the graph structure and the feature matrix $\mathbf{F} \in \R^{|V|\times f}$. During unsupervised learning, the objective is to learn the node embedding matrix $\M \in \R^{|V|\times d}$, where the embedding dimension for each node is $d$. 

\subsection{GNN Training}

Mini-batch processing in GNN training can typically be divided into three main steps: (1) Sampling a subgraph from the original graph. The result is the vertex and edge IDs of the sampled nodes and edges, respectively. (2) Feature fetching corresponding to vertex/edge IDs of the sampled nodes/edges in the sub-graph, and (3) Training the GNN for that mini-batch using the sub-graph topology, and the features and labels for corresponding nodes and edges.

Current distributed training frameworks for GNNs suffer from data communication bottlenecks. Both GraphLearn (GL) \footnote{ByteGNN \cite{bytegnn} is a further optimized version of GraphLearn, where sampling and feature fetching are phases of GNN training are parallelized, and the k-hop neighbor fetching is pipelined.} \cite{aligraph} and DistDGL \cite{dist_dgl}, which are prominent distributed GNN training systems, exhibit a shared architectural approach in feature fetching. The process involves issuing requests from the graph backend to retrieve features for sampled vertices in each mini-batch. This methodology is inefficient for two primary reasons. Firstly, it generates a high volume of feature-fetching requests for each batch, potentially causing high network traffic. Secondly, due to the graph's power-law distribution, there is a high probability of duplicated vertices in neighboring hops. However, the system often processes features for these repeated vertices multiple times, resulting in increased network load. This is illustrated in Fig. \ref{fig:gnn_comm_bottleneck}. 

\subsection{Node embedding training}

Methods such as skip-gram and negative sampling are popular for training graph node embeddings due to their ability to effectively capture complex graph structures and scale to large networks \cite{skipgram1, skipgram2, deepwalk, node2vec}. 
These methods translate graph topologies into a learnable format by simulating random walks, enabling them to handle various types of relationships and interactions between nodes in a computationally efficient manner.
A few notable algorithms for training node embedding algorithms are DeepWalk \cite{deepwalk}, Node2vec \cite{node2vec} and LINE \cite{line}. DeepWalk trains node embeddings by generating random walks from each node in the graph and applying the skip-gram model from NLP for context-based training. Node2vec is an extension of DeepWalk that introduces a flexible, biased random walk to balance between breadth-first and depth-first traversals, allowing for more nuanced exploration and representation of graph structures. In LINE, the objective function utilizes information from both the local and global graph structures around a node to learn its embedding. In this paper, we use DeepWalk and LINE as representative node embedding algorithms to illustrate training with GraphScale. 

As illustrated in Fig. \ref{fig:graph_dp}, there are two bottlenecks to data-parallel training of node embeddings for large graphs, namely storage (to store the model, that is, the embeddings matrix $\M \in \R^{|V|\times d}$ in memory) and communication (to average gradients during each mini-batch update to the model).

\section{GraphScale for Learning on Graphs}
\label{sec:graphscale}
{\bf GraphScale system design}. In GraphScale, our reimagined system architecture for GNN training capitalizes on the separation of storage and computation. 
Throughout this paper, computation workers are referred to as trainers, and workers who store the data are referred to as actors (following Ray's terminology). Note that a single machine can have tens or hundreds of CPUs/cores. Within the Ray framework, each of these cores is capable of functioning either as an actor or a trainer. Despite actors and trainers operating on the same machine, Ray's abstraction permits us to treat storage and computation as distinct entities. This approach not only maximizes efficient usage of resources (such as CPU memory and computational power) but also simplifies resource management for the user by abstracting away the intricacies of actual resource allocation. GraphScale is currently implemented in CPU-only environments, where we emphasize scalability in settings where GPUs are not readily available.


\begin{figure}[t]
\centering
\includegraphics[width=0.3\textwidth]{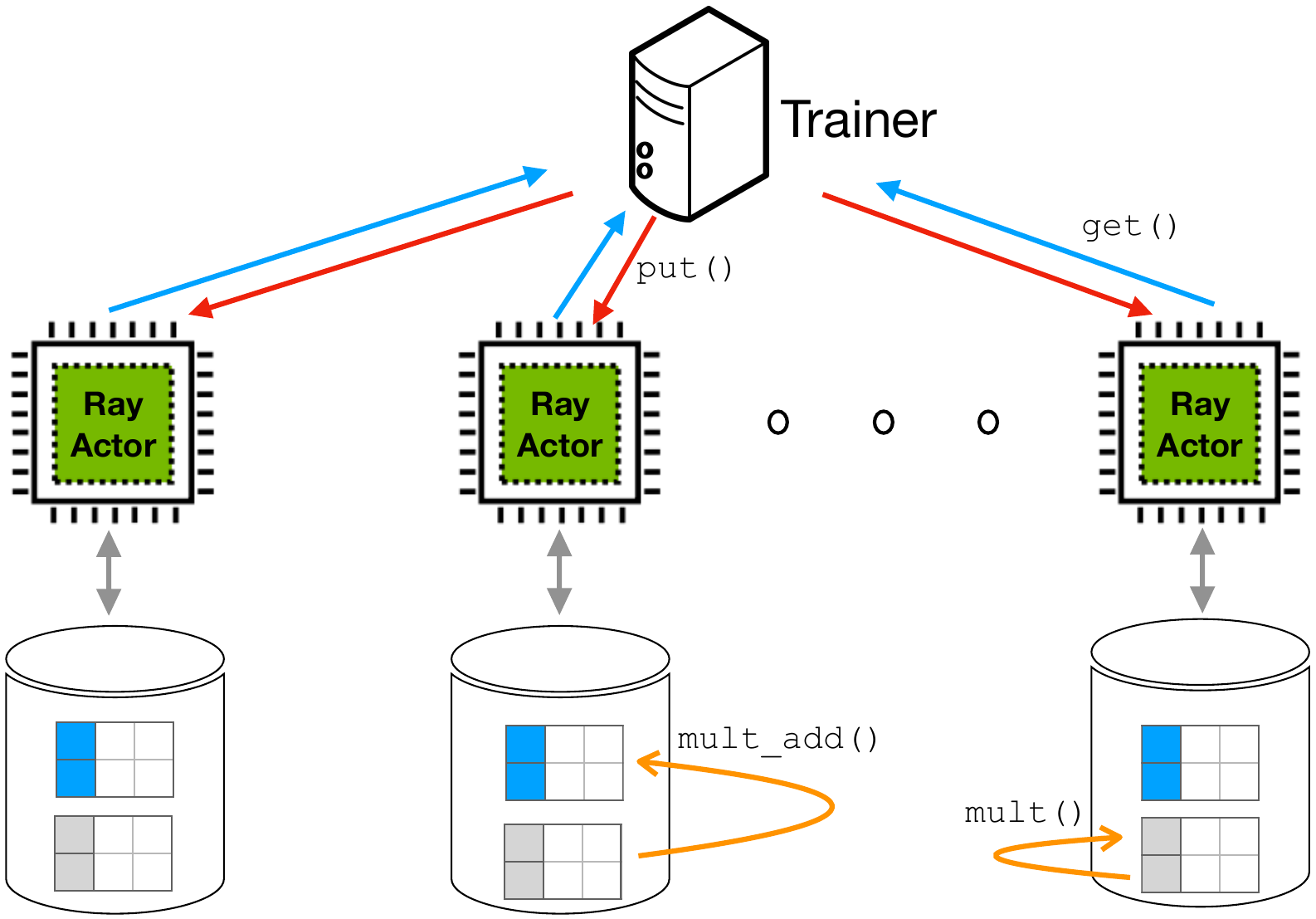}
\caption{\footnotesize An illustration of the GraphScale system, where each Ray actor stores one shard each of two large matrices (such as the embedding and momentum data) in two separate GraphScale data structures. Trainers can use {\tt get()} ({\tt put()}) to read (write) data from (to) the actors. Operations {\tt mult()},  and {\tt mult\_add()} are performed locally in actors' memory and in a parallel fashion.}
\label{fig:ros_operations}
\end{figure}

GraphScale system is illustrated in Fig. \ref{fig:ros_operations}. Separating storage actors and trainers drastically cuts down communication overhead, as it necessitates only a single feature request for the sub-graph from each trainer. This efficiency is largely enabled by Ray's capacity to facilitate an easy abstraction of storage and computation in serverless systems.
GraphScale storage actors also enable effective data updates by allowing operations like read, update, addition, and multiplication as illustrated in Fig. \ref{fig:ros_operations}.

\textbf{Implementation}. GraphScale is implemented using the Ray actor model\footnote{https://docs.ray.io/en/latest/ray-core/actors.html}. The system comprises 2,800 lines of Python code and 500 lines of C++ code. This implementation takes advantage of Ray's flexible scheduling features, allowing for straightforward integration of asynchronous operations and pipeline processing. The C++ component is specifically utilized to enhance the efficiency of indexing and constructing large data arrays.

\begin{figure*}[t]
\centering
\includegraphics[width=0.95\textwidth]{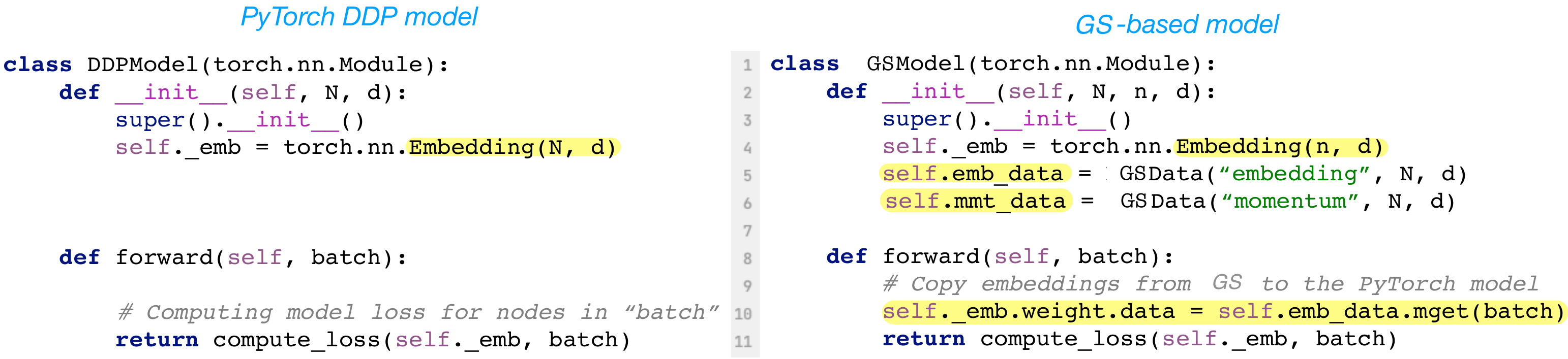}
\vspace{-2mm}
\caption{\footnotesize Comparison of Python codes for PyTorch DDP versus GraphScale (GS) model definitions using PyTorch.}
\label{fig:model_codes}
\end{figure*}

 \begin{figure*}[t]
\centering
\includegraphics[width=0.95\textwidth]{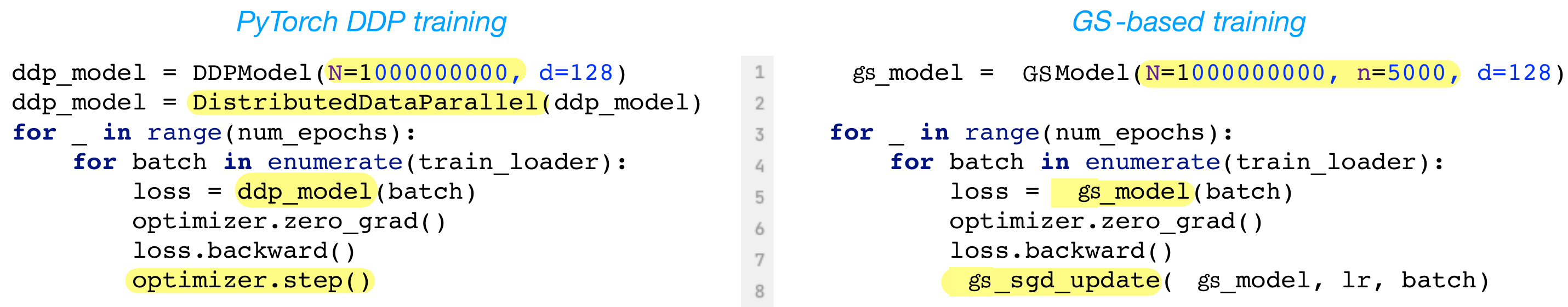}
\vspace{-2mm}
\caption{\footnotesize Comparison of Python codes for DDP versus GraphScale (GS) based graph node embedding training for PyTorch.}
\label{fig:training_codes}
\end{figure*}

\subsection{GNN Training} 

{\bf Reducing feature requests by trainers}. In GraphScale, the trainers keep the entire graph topology\footnote{Note that storing the topology locally in GraphScale can quickly become a bottleneck for large graphs. However, graph compression methods like GBBS \cite{gbbs} can be used to mitigate this bottleneck, as shown by the authors in \cite{lightne}, where they store a 74 billion node graph's topology in just 107 GB of memory.}, but the heavy feature matrix is stored at the storage actors. To use network bandwidth efficiently during GNN training, the trainers combine multiple feature requests into a single batch at the end of each mini-batch training session. By keeping the entire graph topology, trainers can merge the same vertex IDs from subgraphs before making (a single) feature request to the actors. The actors then respond with one data block per request. 

This method of consolidating requests and removing duplicate vertices greatly reduces the network bandwidth needed. It also reduces the dependence on the partitioning scheme, since regardless of the partition, only one communication request is needed per trainer. However, the size of the data to be communicated still depends on the partitioning scheme, but its effect on the overall training time is negligible. In Sec. \ref{sec:ros_exps_gnn}, we show that training GNNs with GraphScale is at least $30\%$ faster than distributed GNN training frameworks like DistDGL \cite{dist_dgl} and GraphLearn (also known as AliGraph) \cite{aligraph}. Next, we describe node embedding training with GraphScale.

\subsection{Node Embedding Training}
\label{sec:gs_emb_training}

Like GNN training, GraphScale utilizes its storage actors to scale the embedding matrix with the size of the graph, mitigating the storage bottleneck in data-parallel training of the node embedding matrix. To mitigate the communication bottleneck, we exploit the fact that gradients are sparse, as observed below.

During training a mini-batch with $B$ source nodes, let $n$ be the maximum number of nodes obtained during graph sampling (including source nodes in mini-batch and their positively and negatively sampled nodes). 
The number $n$ is typically easy to estimate, e.g., for DeepWalk, it is $B \times $ {\tt walk\_len}, where {\tt walk\_len} is the length of the random walk used for DeepWalk training, and for LINE, it is $B \times (2 + {\tt num\_neg})$, where {\tt num\_neg} is the number of negatively sampled nodes per source node in the mini-batch. For example, a good batch-size and walk length for DeepWalk is $B = 1000$ and {\tt walk\_len=5}, respectively.
We generally have $N \gg n$, that is, the total number of nodes in the graph (e.g., $N=1~billion$) is much greater than the number of nodes sampled in a mini-batch (e.g., $n=5000$). Thus, the gradients during node embedding training are typically sparse. This allows each trainer in GraphScale to communicate only an update of size $n$ instead of size $N$ in data-parallel training to update the model in each iteration as we describe next. 

PyTorch's DistributedDataParallel (DDP) is a widely used method for distributed data-parallel training of large models and is also used in graph frameworks like GraphLearn (GL) \cite{aligraph}. We use DDP training as a baseline to illustrate node embedding training with GraphScale. Figures \ref{fig:model_codes} and \ref{fig:training_codes} detail the Python codes for model definition and training loops, highlighting implementation differences in yellow. Both methods are implemented on PyTorch.

Fig. \ref{fig:model_codes} shows the common aspects of model definitions shared by DDP and GraphScale. In line 4, the PyTorch embeddings are initialized (however, with different sizes). In lines 5 and 6 in the GraphScale case, we initialize the GraphScale storage for node embeddings and momentum vector, respectively, where the dimension of GraphScale matrices is $N \times d$. This initializes the embeddings and momentum matrices distributedly using Ray. In line 11, {\tt compute\_loss} is a loss function that computes the loss using the current batch (which includes the source and positively and negatively sampled nodes). 

\begin{figure}[t]
\centering
\includegraphics[width=0.44\textwidth]{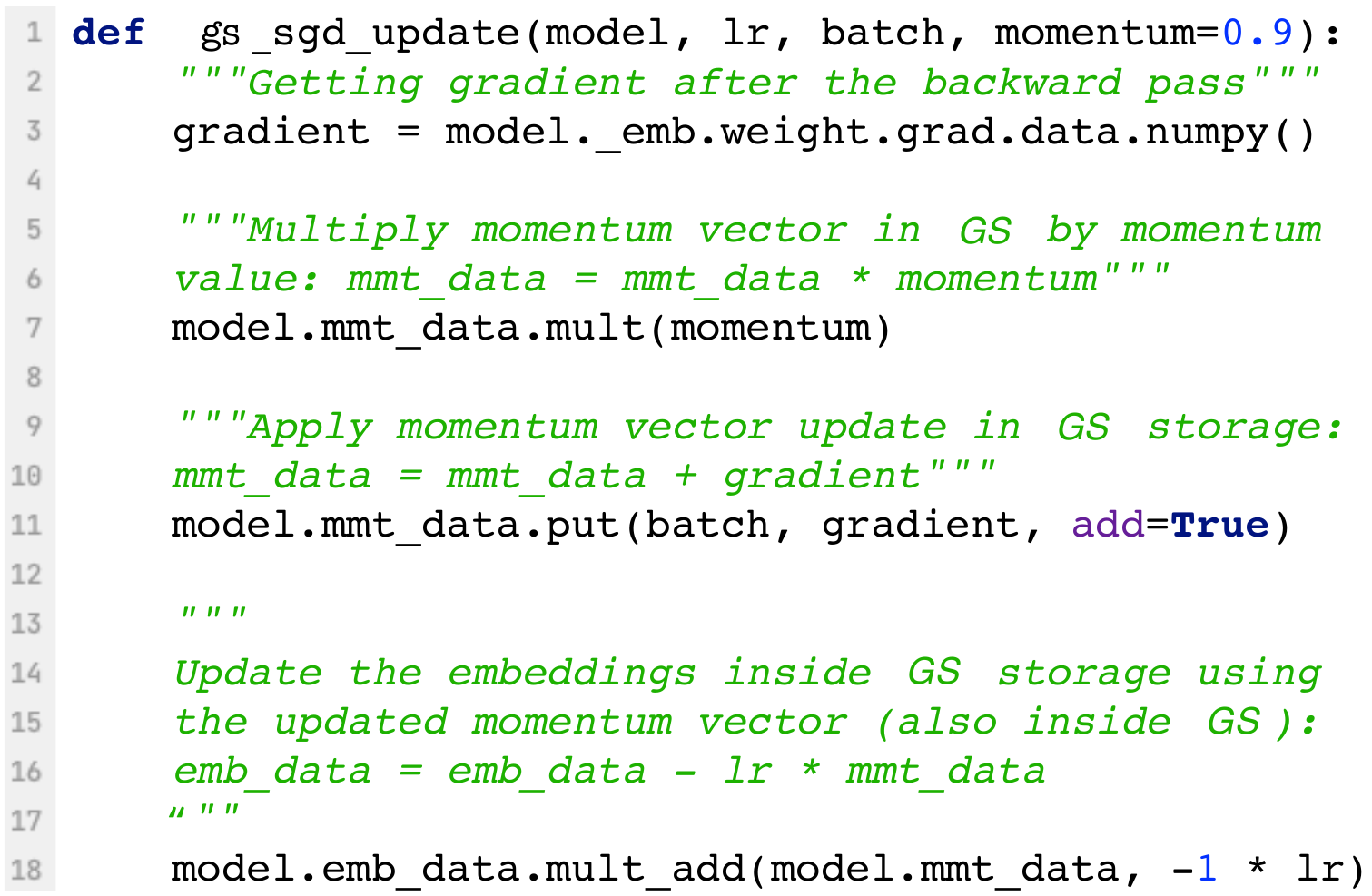}
\caption{\footnotesize Python code for SGD with momentum update when both the embeddings and momentum vectors are stored in GraphScale (GS) storage.}
\label{fig:ros_momentum_update}
\end{figure}

Fig. \ref{fig:training_codes} describes the pseudo-code for DDP and GraphScale-based PyTorch training loop. Note that the implementation for GraphScale-based training is as straightforward as only a few lines of code changes. The most notable points are:
\begin{itemize}[leftmargin=*]
\setlength{\parskip}{0pt}
\setlength{\itemsep}{0pt}
    \item GraphScale requires a significantly smaller memory at each worker to hold the model. The size of the embedding matrix in {\tt ddp\_model} is $N \times d$ (Line 4, Fig. \ref{fig:model_codes}), where $N = 1$ billion (Line 1, Fig. \ref{fig:training_codes}). In comparison, the {\tt gs\_model} in GraphScale has an embedding matrix of dimension $n \times d$, where $N \gg n$, e.g., $n = 5000$ when the mini-batch size, $B = 1000$, and the walk length is 5 in DeepWalk.
    \item In GraphScale, workers communicate smaller gradients after each iteration. DDP uses {\tt ddp\_model} to wrap PyTorch's DistributedDataParallel class for data-parallel training (Line 2, Fig. \ref{fig:training_codes}). This means gradients are averaged before model updates, communicating an $N\times d$ matrix in DDP, compared to $n \times d$ in GraphScale.
    \item In GraphScale, each worker fetches only a subset of the model for each iteration. The loss (and the gradient) is computed using the sampled node embeddings for the local mini-batch at each worker and not all the nodes (Line 11, Fig. \ref{fig:model_codes}). Thus, each worker needs embeddings of $n$ nodes instead of $N$ nodes from the storage to compute loss on its local subgraph.
    \item GraphScale doesn't rely on a central parameter server for momentum and model updates by distributing them within its storage. Hence, instead of using PyTorch's optimizer, we implement custom optimizers like SGD with momentum \cite{sgd_momentum} and Adam \cite{adam} within GraphScale. This is illustrated in Fig. \ref{fig:training_codes} (Line 8) and Fig. \ref{fig:ros_momentum_update}.
\end{itemize}

In Fig. \ref{fig:ros_momentum_update}, we describe our implementation of model update in GraphScale for SGD with momentum. In Line 3, we obtain the gradient of the embedding vectors for nodes that are a part of the subgraph created by the corresponding mini-batch. In Line 7, we multiply the momentum data in GraphScale by the momentum value using the GraphScale {\tt mult()} operation. In Line 11, we add the gradient corresponding to the nodes in the batch to update the momentum vector for that iteration using the GraphScale {\tt put()} operation. Finally, in Line 18, the Ray actors perform an SGD update of the model by first multiplying the momentum data by the learning rate and then subtracting it from the current model data using the {\tt mult\_add()} operation. 
Note that operations such as {\tt mult()}, {\tt mult\_add()}, {\tt get()} and {\tt put()} operations are completed by multiple Ray storage actors parallelly that store data like model and momentum and do not require any raw data communication (see Fig. \ref{fig:ros_operations} for an illustration). 

\textbf{Synchronization}. There are two types of synchronizations required: across iterations and across trainers. 
To achieve synchronization across iterations, we add a barrier after the write operation at the end of each iteration (that is, after the trainers update embedding and momentum). This barrier allows trainers to read updated values for the next iteration. 
During the write operation, synchronization is also required among parallel trainers (as they might be updating embeddings for the same nodes). However, we do not apply strict synchronization between the trainers and allow them to write incremental updates to embeddings of the same nodes. This is because such asynchronous methods are faster in distributed settings due to no locking and have the same theoretical rate of convergence as synchronous SGD \cite{hogwild, async_sgd}.
Further, we evaluate the convergence of training with GraphScale in Section \ref{sec:ros_exps_emb} and show that it performs on par with synchronized SGD while enjoying at least 70\% speedup.

\section{Experimental Results}
\label{sec:ros_exps}

To evaluate GraphScale's performance, we divide our experiments into two parts: GNN algorithms and node embedding algorithms, respectively, in Section \ref{sec:ros_exps_gnn} and Section \ref{sec:ros_exps_emb}.  For GNN training, we compare GraphScale against two popular distributed frameworks, GraphLearn (GL) \cite{aligraph} and DistDGL \cite{dist_dgl},
in terms of throughput and latency with GraphSage \cite{graphsage} as a representative algorithm. For node embedding training, we compare GraphScale with PyTorch DDP and PyTorch BigGraph \cite{pbg_biggraph} with DeepWalk \cite{deepwalk} and LINE \cite{deepwalk} as representative algorithms. 

{\bf Experimental setup}: Each machine has 48 physical cores and 700 Gigabytes of memory. As a result, we get 48 actors per machine. The distributed cluster consists of 4 such machines.
The ethernet provides a communication bandwidth of 25 gigabits per second. 

{\bf Datasets}: From publicly available datasets, we use Reddit \cite{graphsage}, ogbn-products \cite{ogb} and ogbn-papers100M \cite{ogb}. Reddit is a graph dataset extracted from Reddit posts made in the month of September 2014 with 232,965 nodes as posts and a feature dimension of 602. ogbn-products is an undirected graph that has approximately 2.5 million nodes and 62 million edges and represents an Amazon product co-purchasing network, while ogbn-papers100M has 111 million nodes and 1.6 billion edges and is a directed graph representing a paper citation network. Finally, we also implement GraphScale-based node embedding on a proprietary dataset from the TikTok social media platform with 1 billion nodes and 92 billion edges.

{\bf Configuration}:
Unless mentioned otherwise, we use the following as the default configurations. The batch size is 512. The number of epochs is 4. The number of trainers per machine is 2. The learning rate is ($0.003$) for GraphSage, ($0.01$) for DeepWalk, and ($1.0$) for LINE. The default sampling fanout for GraphSage is [15, 10]. We use a walk-length of 5 and a window-size of 3 for DeepWalk; and 5 negative nodes per source node for LINE.



\subsection{Training GNNs with GraphScale}
\label{sec:ros_exps_gnn}
Experiments in this section compare the throughput of feature fetching, assessing its impact on latency, and comparing the overall performance of GraphScale with DistDGL and GL across all three stages--sampling, feature fetching, and training on the Reddit graph dataset.

We define throughput as the average number of features fetched per second.
We vary two variables to evaluate GraphScale's scalability and throughput: the number of trainers and sampling fanout.
Trainers are the computation workers doing feature fetching in parallel, while fanout represents the aggregation workload per batch. High fanout leads to higher accuracy but increased workload since more neighbors need to be aggregated. 



\begin{figure}[t]
        \centering

    \begin{subfigure}[t]{0.235\textwidth}
        \centering
        \includegraphics[scale=0.36]{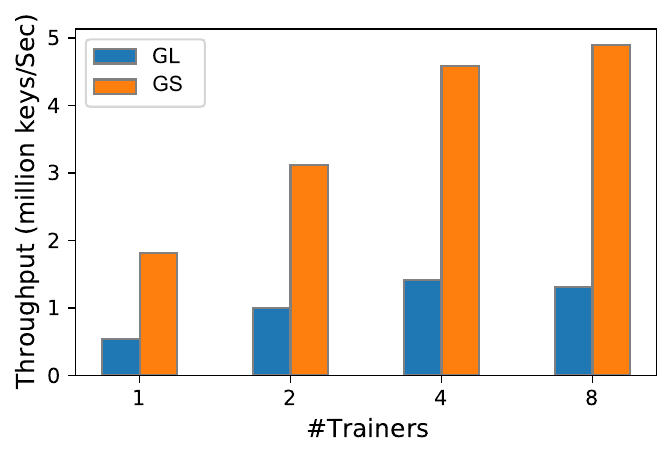}
    \end{subfigure}
    \begin{subfigure}[t]{0.235\textwidth}
        \centering
        \includegraphics[scale=0.36]{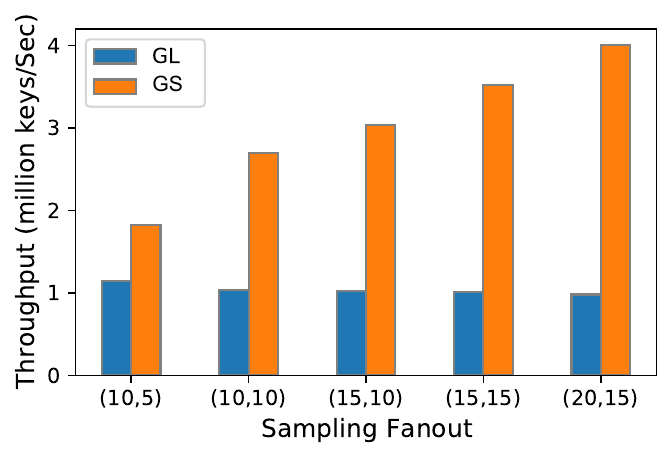}
    \end{subfigure}
\vspace{-5mm}
\caption{\footnotesize Throughput (in million vertices per second) for GraphSage algorithm with the Reddit dataset. GraphScale (GS) increases the throughput up to $4\times$ when compared to GL.}
\label{fig:th_single}
\end{figure}

Fig. \ref{fig:th_single} presents a comparison with GL on a cluster of 4 machines.
With the number of trainers per machine increasing from 1 to 8, GraphScale has a steady growth while GL has a small throughput increase. With 8 trainers per machine, GraphScale's throughput is $3.73\times$ of GL's.
Next, we vary the sampling fanout from [10, 5] to [20, 15] (that is, now we aggregate 20 and 15 neighbors in the first and second hop, respectively). Unlike GL, the throughput increases commensurately with fanout (that is, computation load) for GraphScale. For the fanout of [20, 15], GraphScale has a throughput of $4\times$ the throughput of GL. 
This is because GraphScale reduces communication time during feature fetching and mitigates the I/O bottlenecks to utilize computational resources more efficiently.

\begin{figure}[t]
    \centering
    \begin{subfigure}[t]{0.235\textwidth}
        \centering
        \includegraphics[scale=0.36]{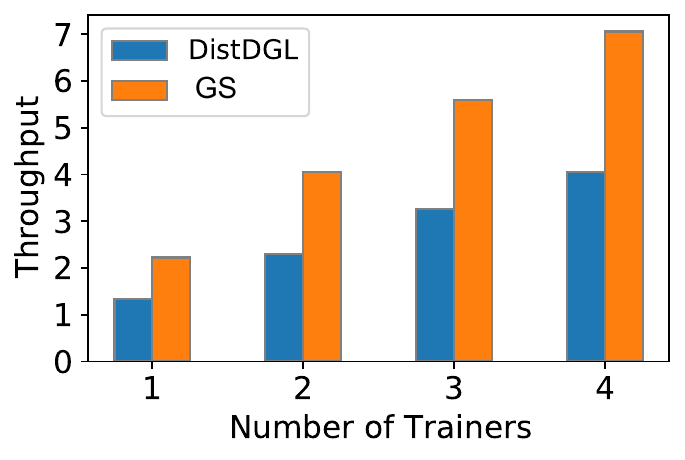}
    \end{subfigure}
    \begin{subfigure}[t]{0.235\textwidth}
        \centering
        \includegraphics[scale=0.36]{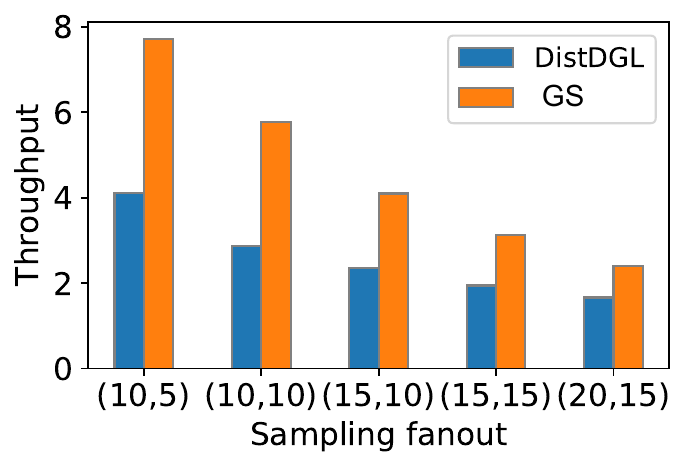}
    \end{subfigure}
\vspace{-6mm}
\caption{\footnotesize Throughput comparison between DistDGL and GraphScale (GS) on GraphSage with the Reddit dataset. GraphScale has significantly better throughput than DistDGL in all scenarios.}
\label{fig:ros_e2e_perf}
\end{figure}

Similarly, Fig. \ref{fig:ros_e2e_perf} compares the performance of GraphScale with DGL.
When the number of trainers increases, the throughput of both DistDGL and GraphScale grows proportionally. On average, GraphScale increases the training throughput by $71.7\%$.
When the sampling fanout increases, both DistDGL and GraphScale have reduced throughput because both feature-fetching and training time increase due to the growing workload. On average, GraphScale improves DistDGL's throughput by 73.8\%.

We also evaluated GraphScale on the ogbn-papers100M dataset and observed similar trends. Specifically, GraphScale achieved at least a $1.4\times$ higher throughput compared to GL and DistDGL. Detailed results are omitted due to space limitations.


\subsection{Node Embedding Training with GraphScale}
\label{sec:ros_exps_emb}

In this section, we evaluate the performance of GraphScale on training node embeddings of graphs through experiments on several real-world datasets. 
Throughout the node embedding experiments, we use a Ray cluster of 4 machines and eight trainers while training. We perform our experiments on two popular node embedding algorithms, DeepWalk and LINE, to show the efficacy of GraphScale-based node embedding training.


\begin{figure*}[t]
    \centering
    \begin{subfigure}[t]{0.235\textwidth}
        \centering
        \includegraphics[scale=0.5]{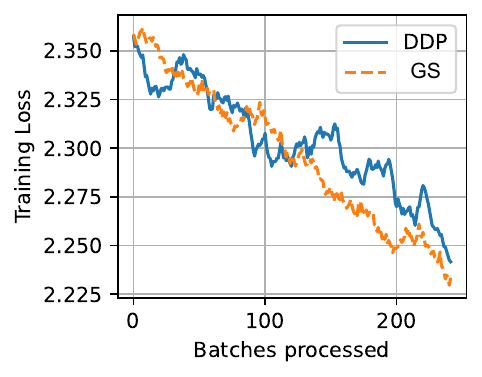}
        \caption{Training loss for DeepWalk.}
        \label{fig:128_papers_deepwalk_iter}
    \end{subfigure}
    ~
    \begin{subfigure}[t]{0.235\textwidth}
        \centering
        \includegraphics[scale=0.23]{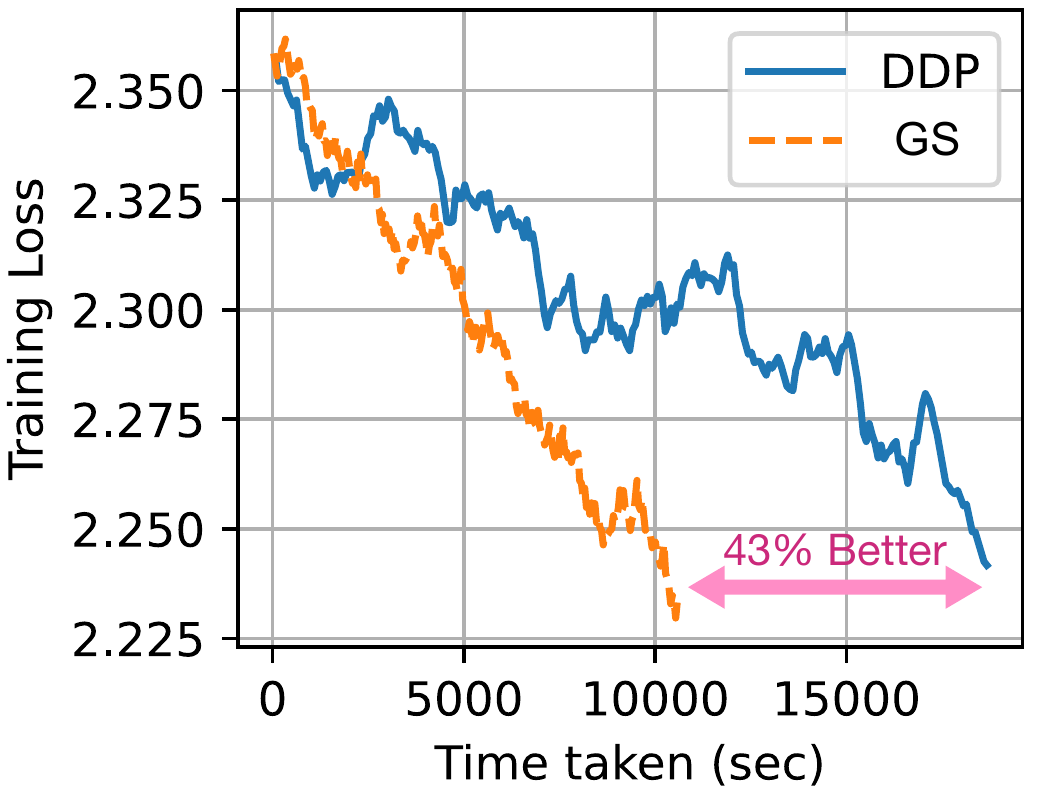}
        \caption{Training time for DeepWalk.}
        \label{fig:128_papers_deepwalk_time}
    \end{subfigure}
        \begin{subfigure}[t]{0.235\textwidth}
        \centering
        \includegraphics[scale=0.5]{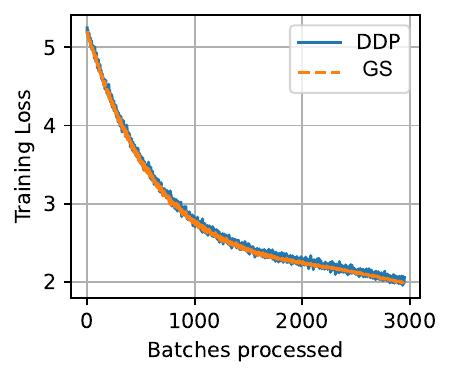}
        \caption{Training loss for LINE.}
        \label{fig:128_papers_line_iter}
    \end{subfigure}
    ~
    \begin{subfigure}[t]{0.235\textwidth}
        \centering
        \includegraphics[scale=0.23]{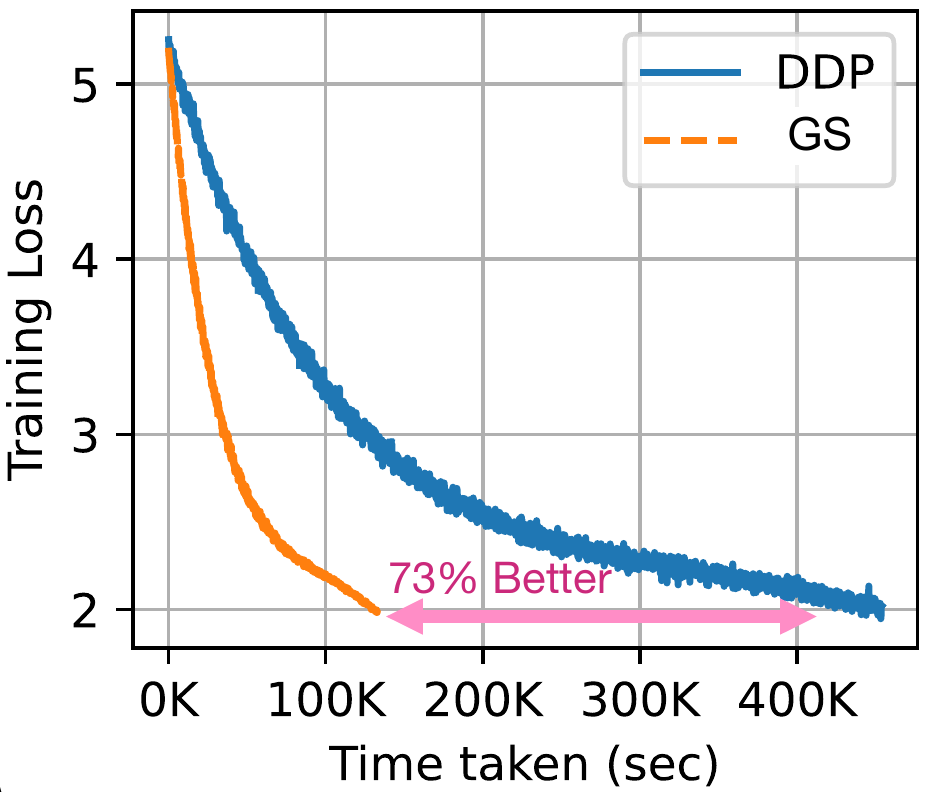}
        \caption{Training time for LINE.}
        \label{fig:128_papers_line_time}
    \end{subfigure}
    \vspace{-3mm}
    \caption{\footnotesize Training stats with the ogbn-papers graph. GraphScale (GS) has the same accuracy as DDP but takes at least $43\%$ less time.}
\label{fig:papers_128}
\end{figure*} 

{\bf GraphScale outperforms data-parallel training}: First, we compare GraphScale with data-parallel training (that is, PyTorch DDP) which is employed in popular frameworks like GL and PyTorch Geometric. In Fig. \ref{fig:papers_128}, we plot the training loss for GraphScale and DDP with respect to the number of batches processed and the amount of time taken for the ogbn-papers100M datasets, with an embedding size of 32. 
The training statistics were averaged over three independent trials. We highlight two key observations below.\\
1. {\bf Training Loss:} GraphScale has similar (if not better) training accuracy when compared with DDP. This is evident from figures \ref{fig:128_papers_deepwalk_iter} and \ref{fig:128_papers_line_iter} where we plot the smoothened training loss for DeepWalk and LINE, respectively.\\
2. {\bf Training Runtime}: GraphScale is significantly faster than DDP. This is evident from figures 
\ref{fig:128_papers_deepwalk_time} and \ref{fig:128_papers_line_time}, where we observe savings of $43\%$ for DeepWalk and $73\%$ for LINE, respectively.

We also performed the same experiments on the ogbn-products datasets for both DeepWalk and LINE, and observed savings of $39\%$ and $68\%$, respectively, without any loss in performance.

\begin{figure*}[!htb]
\minipage{0.32\textwidth}
\centering
\includegraphics[scale=0.43]{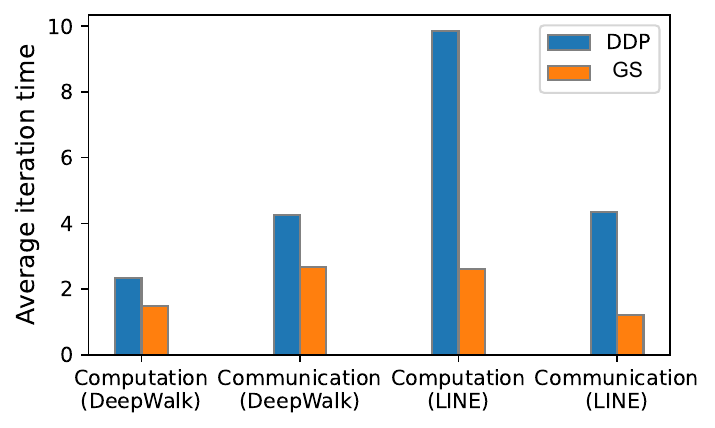}
\caption{\footnotesize Runtime comparison for computation and communication phases with GraphScale and DDP for DeepWalk and LINE on the ogbn-products dataset (with embedding size 128). We observe commensurate savings with computation and communication.}
\label{fig:bar_plot_comp_comm}
\endminipage\hfill
~
\hspace{0.15cm}
\minipage{0.32\textwidth}
\centering
\includegraphics[scale=0.2]{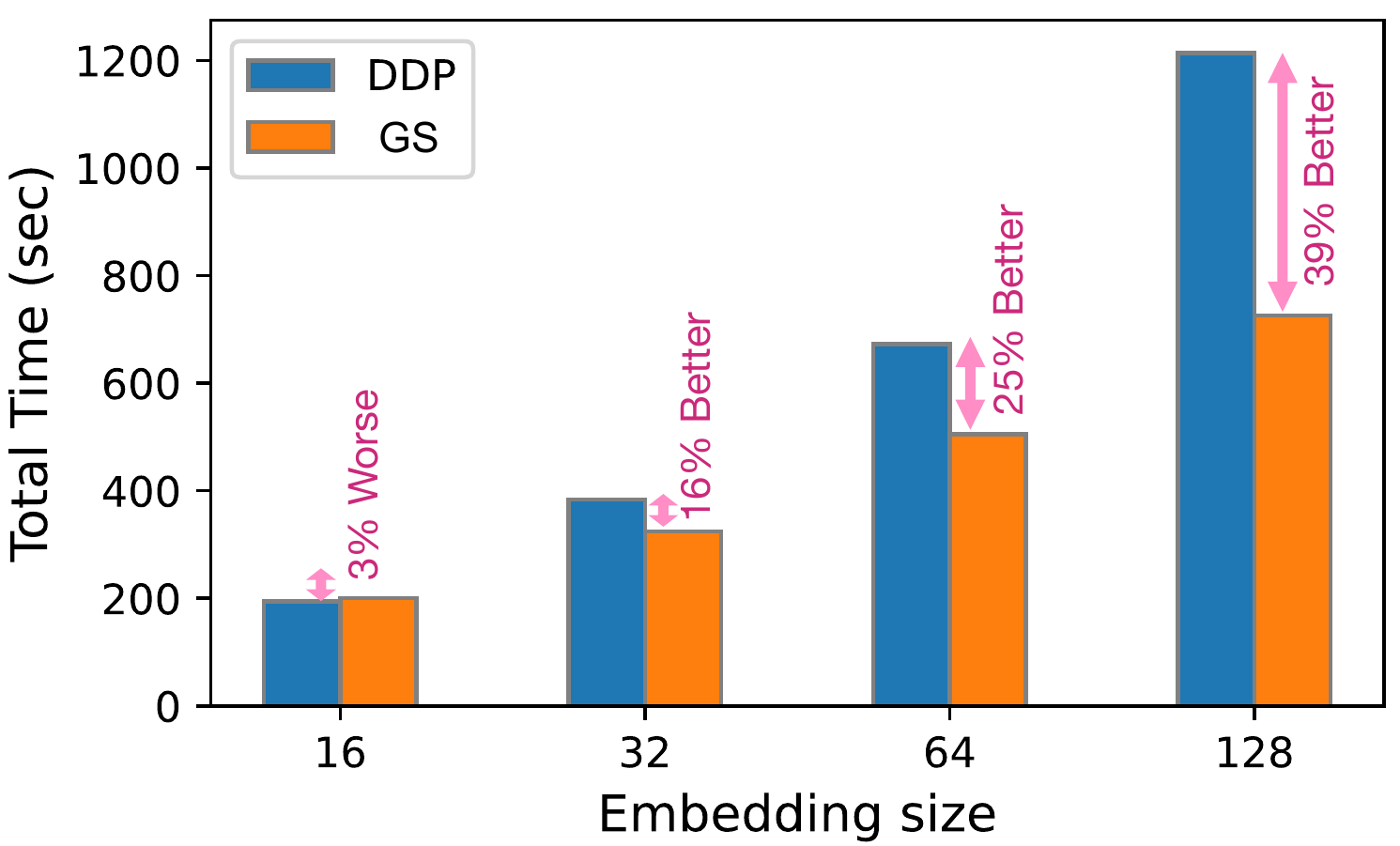}
\caption{\footnotesize Total runtime comparison for different embedding sizes training DeepWalk on the ogbn-products dataset. Savings with GraphScale increase with embedding size due to less communication and computation when compared to DDP.}
\label{fig:bar_plot_diff_emb_sizes}
\endminipage\hfill
~
\hspace{0.1cm}
\minipage{0.30\textwidth}
\centering
\includegraphics[scale=0.43]{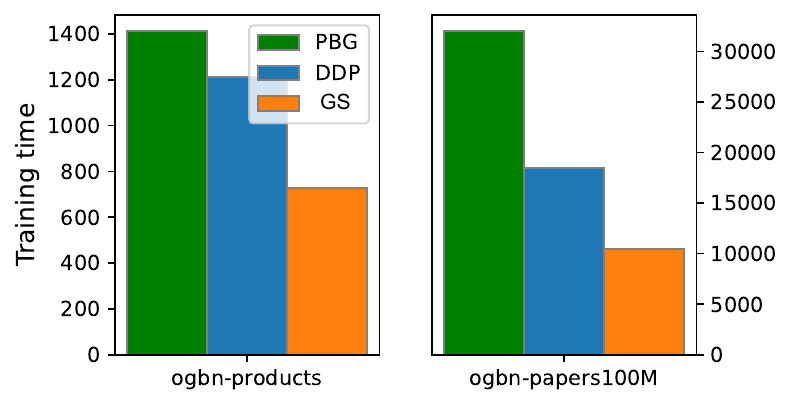}
\caption{\footnotesize Total runtimes for PBG, DDP and GraphScale on the two datasets for DeepWalk to achieve the same training loss. PBG is worse than both DDP and GraphScale due to slower convergence and under-utilization of resources.}
\label{fig:bar_plot_pbg}
\endminipage\hfill
\vspace{-3mm}
\end{figure*}

{\bf GraphScale saves both communication and computation time}: In Fig. \ref{fig:bar_plot_comp_comm}, we compare the average computation and communication time per iteration for GraphScale with DDP. Computation time primarily comprises of the forward and backward passes, and communication time comprises the optimizer and model updates (that is, allreduce in the case of DDP and GraphScale storage updates in the case of GraphScale). We note that GraphScale outperforms DDP significantly during both communication and computation. During communication, for large embedding sizes, the gradients become large too, and PyTorch DDP requires a lot of communication during allreduce. On the other hand, by virtue of only training nodes in the mini-batch and its sampled neighbors (that is, the subgraph for that mini-batch), GraphScale reduces the amount of communication required. During computation, GraphScale further takes less time since it performs operations (such as forward and backward passes) on a PyTorch tensor of the size of the subgraph, while DDP performs these operations on the entire embedding matrix. On a side note, we can also see that DeepWalk spends more time in communication, while LINE requires more time in computation.

{\bf GraphScale is scalable}: In Fig. \ref{fig:bar_plot_diff_emb_sizes}, we plot the total runtimes for the ogbn-products dataset with DDP and GraphScale for different embedding sizes and observe that GraphScale performs better as the embeddings get larger. GraphScale is slower for small embedding sizes since it requires two rounds of communication in each iteration, one for copying the current embedding values from GraphScale to memory and the second for updating the model in GraphScale after training. On the other hand, PyTorch DDP requires only one round of communication when the gradients are averaged. However, DDP is not scalable since it requires both higher communication and computation as it operates on the entire DDP matrix (as shown in Fig. \ref{fig:bar_plot_comp_comm} and explained earlier).

{\bf Comparison with PBG}: PyTorch-BigGraph (PBG) \cite{pbg_biggraph} addresses the memory bottleneck in graph node embedding training by partitioning nodes into $M$ buckets (where $M$ is the number of machines). This divides the edges into $M^2$ buckets depending on the source and destination nodes. Thus, it enables large-scale embedding training when embeddings don't fit in memory by limiting training to disjoint edge buckets. PBG's architecture involves a lock server, parameter server, partition server, and shared filesystem, which can be complex to implement in production environments. 
In Figure \ref{fig:bar_plot_pbg}, we compare PBG, DDP, and GraphScale in reaching a training loss of 4.45 with the same initial model values.
PBG takes significantly longer than DDP and GraphScale due to several factors. PBG's positive and negative sampling is not independent and identically distributed as it samples only from the local bucket. Furthermore, PBG's restriction that only one machine can work on one node partition makes training not embarrassingly parallel, leading to underutilization of resources.



{\bf Industry-scale datasets}: We trained DeepWalk on an industry-scale dataset featuring 1 billion nodes and 92 billion edges, using SGD with momentum (learning rate: 0.01, batch size: 2048, momentum: 0.9). Distributed training utilized a Ray cluster of 10 machines, each with 48 cores and 700 GB of memory, totaling 480 Ray actors serving both compute and storage. With an embedding size of 128, the size of the node embedding matrix becomes 256 GB in fp16 precision. Storing it along with the graph topology (and the momentum vector) at each node makes PyTorch DDP training infeasible. Thus, we employ GraphScale, which requires that each storage actor holds one-tenth of the embedding matrix (25.6 GB in this case). We observe that a mini-batch of size 512 takes 0.8 seconds on average, and training converges to a loss of 0.4 in slightly less than 3 hours. 

GraphScale is currently deployed at TikTok and its Chinese counterpart Douyin, two of the world's largest social media platforms, for both supervised and unsupervised graph learning at scale. 

\section{Related Work}
\label{sec:related_work}


Many single-machine systems have been proposed in the literature to tackle the sampling bottleneck in GNN training. PyTorch-Geometric \cite{pyg} implements a message-passing API for GNN training. Using the Apache TVM compiler, FeatGraph \cite{hu2020featgraph} generates optimized kernels for GNN operators for both CPU and GPU. PaGraph \cite{pagraph} proposes a GPU caching policy to address the subgraph data loading bottleneck. NextDoor \cite{nextdoor} proposes a graph-sampling approach called transit-parallelism for load balancing and caching of edges. \cite{gns_da} propose Global Neighborhood Sampling (GNS) for mixed CPU-GPU training, where a cache of nodes is kept in the GPU through importance sampling, allowing for the in-GPU formation of mini-batches. 
\cite{mlsys_gnn_sampling} uses a performance-engineered neighborhood sampler to mitigate mini-batch preparation and transfer bottlenecks in GNN training.
We note that many of the above schemes for faster sampling for GNN training can be used complementarily with GraphScale.

The above single-machine systems are limited by CPU/GPU memory when processing large industrial-scale graphs.
For GNN training on such large graphs, distributed training was utilized in DistDGL \cite{dist_dgl} and GraphLearn \cite{aligraph}. However, it introduces a key performance bottleneck: high sampling time due to random data access and remote feature fetching. As noted from our experiments in Sec \ref{sec:ros_exps_gnn}, GraphScale is able to mitigate this bottleneck by separating the topology and feature storage in graphs. 


Despite its industrial importance, research on node-embedding training for large graphs remains nascent. Authors in \cite{nimble_tt_decomp} propose tensor-train decomposition \cite{tt-decomp} to compress embeddings by enabling their compact parametrization. This method can easily be incorporated into node embedding training with GraphScale to further reduce memory and communication bottlenecks. Marius \cite{marius} was proposed for a single machine GPU training of node embeddings to reduce the data movement by leveraging partition-caching and buffer-aware data orderings. Moreover, techniques like gradient quantization and sparsification \cite{Stich2018Sparsified, dct_kdd} can be applied complementarily to GraphScale to reduce communication costs.
PBG \cite{pbg_biggraph} proposed a distributed solution but complicated partition-aware distributed training that is not embarrassingly parallel. Also, it requires a large shared parameter server to store the node embeddings while training. 
As we observed in our experiments in Sec. \ref{sec:ros_exps_emb}, GraphScale outperforms PBG due to a better parallelization strategy that utilizes the sub-graph structure while storing the node embeddings distributedly.
\newpage
\bibliographystyle{ACM-Reference-Format}
\balance
\bibliography{bibli}

\end{document}